\documentclass{article}
\usepackage{neurodata}

\title{Geodesic Learning via Unsupervised Decision Forests}

\author[1] {Meghana Madhyastha}
\author[2]{Percy Li} 
\author[1] {James Browne}
\author[3]{Veronika Strnadova-Neeley}
\author[2]{Carey E.~Priebe}
\author[1]{Randal Burns}
\author[4]{Joshua T.~Vogelstein\thanks{\href{mailto:jovo@jhu.edu}{jovo@jhu.edu}}}
\affil[1]{Department of Computer Science, Johns Hopkins University}
\affil[2]{Department of Applied Mathematics and Statistics, Johns Hopkins University}
\affil[3]{Department of Computer Science, Montana State University}
\affil[4]{Department of Biomedical Engineering, Institute for Computational Medicine, Kavli Neuroscience Discovery Institute, Johns Hopkins University}

\begin{document}

\maketitle

% TODO@meghana: share link to code for all experiments with us

% TOD@VSN: everything else :-)

% TODO@VSN: read this https://bitsandbrains.io/2018/09/08/figures.html 
%VSN: Done (read)

%and make sure all figures are compliant.
%VSN: well that'll take some work...probably won't be done by 6/28

% TODO@VSN: additional figures to explore for journal:
% 1. runtime experiment, varying n, d, and # cores
% 2. same as 4 (top), but varying N
% 3. same as 4 (top), but vary @k
% 4. same as 4, but using supervised
% 5. same as 5, but using brain RNAseq data

%TODO@VSN: condsider the following figure:
% the goal is to understand what is the best trade-off for # of leaf nodes as a function of N samples. 
% given N samples, say with 10 noise dimensions, vary # of trees on 1 dimension, vary tree depth on the other.  make a heatmap showing geodesic precision @k. do this for n=100, 500, 1000, 5000 samples, so 4 heatmaps.  for tree depth, vary it on a log 2 scale, eg, 2^p, where p=[1;log2(N)]. maybe same deal for T = # of trees. 
% even better than doing a fixed depth, we could include a convergence parameter for the BIC score.  in particular, we could require BIC(2-means) > h_N, for a sequence of h's.  we could also consider the rule that we only split when BIC(2-means) is better than BIC(1-mean). 

% jovo: update abstract
\noindent \textbf{Geodesic distance is the shortest path between two points in a Riemannian manifold. Manifold learning algorithms, such as Isomap, seek to learn a manifold that preserves geodesic distances. However, such methods operate on the ambient dimensionality, and are therefore fragile to noise dimensions.  We developed an unsupervised random forest method (URerF) to approximately learn  geodesic distances in linear and nonlinear manifolds with noise.  URerF operates on low-dimensional sparse linear combinations of features, rather than the full observed dimensionality.  To choose the optimal split in a computationally efficient fashion, we developed a fast Bayesian Information Criterion statistic for Gaussian mixture models.  
 We introduce geodesic precision-recall curves that   quantify performance relative to the true latent manifold.  Empirical results on simulated and real data demonstrate that URerF is robust to high-dimensional noise, whereas other methods, such as Isomap, UMAP, and FLANN, quickly deteriorate in such settings. In particular, URerF is able to estimate geodesic distances on a real connectome dataset better than other approaches. }

\section{Introduction}
Learning to organize, rank, and sort points from a data corpus is a fundamental challenge in data science, and a statistical primitive underlying many computer science and machine learning tasks.  For example, nearest neighbor algorithms, which sort all points according to their distances to one another, 
are considered among the top 10 most important algorithms of all time~\cite{Wu2007-zc}, and  have strong theoretical guarantees for both  classification and regression~\cite{Stone1977-fi}.  Decision trees (such as CART and C4.5) can also reasonably be thought of as algorithms for organizing data in a hierarchical fashion; these are among the top ten algorithms as well~\cite{Wu2007-zc}. Moreover, decision trees underlie both random forests~\cite{breiman2001random} and gradient boosted trees~\cite{Freund1997-vd}, which are the two leading algorithms for machine learning on tabular data today~\cite{Caruana2006-wp, Caruana2008-tb, Chen2016-fx}. 
% and both decision forests and deep learning have been demonstrated to approximate nearest neighbor learning rules~\cite{Breiman2000-pq,Giryes2015-tp}. 
To complement the above supervised machine learning settings, there is a rich literature on approximate nearest neighbor algorithms (see Aumüller et al.~\cite{Aumuller2017-xm} for benchmark comparisons of many state of the art approaches), which are used extensively in big data systems. 

Operating on the exact nearest neighbors, (or trying to approximate them), is not always desirable.  For example, consider a simplest  supervised learning setting.  Given a data corpus, $\{(x_n,y_n)\}_{n=1}^N$, learn a decision rule such that, given a new data point $x$, its prediction of $y$ has small error with high probability. A canonical approach is kernel regression~\cite{Scholkopf2002-ar}. A kernel machine's prediction is a weighted linear combination of predictions that the neighbors of $x$ would make, specifically, $\hat{y} = \frac{1}{N} \sum_{n=1}^N y_n \times \kappa (x, x_n)$, for some suitably chosen kernel $\kappa$ (for example, a radial basis function or $k$-nearest neighbors kernel). Such approaches enjoy strong theoretical guarantees~\cite{Mohri2018-tf}. Now, further assume that the $x$'s are noisy measurements of some true, but unobserved $\tilde{x}$'s. Such an assumption, called ``measurement error modeling''~\cite{Fuller1987-oa}, could reasonably be argued to be much more accurate than assuming  the $x$'s are {noise-free measurements}~\cite{Hand2016-vc}.  Under this measurement error assumption, a better approach---meaning an approach that likely achieves  smaller error given the same sample size---would be a kernel regression function on the noise-free measurements, $\hat{y}= \frac{1}{N} \sum_{n=1}^N y_n \times \kappa (\tilde{x}, \tilde{x}_n)$.  Unfortunately, because the $\tilde{x}$'s are not observed, such an approach is not available.  This modeling framework therefore motivates {learning the latent structure of the data, even for low-dimensional data}.  In particular, it motivates learning which sample points are close to one another on an underlying latent structure (such as a manifold), for subsequent inference. Moreover, even though the above task is a supervised learning task, performance improves when learning the structure of only the features $x$, while not even considering the labels $y$ .

Learning latent structure is even more important in the ``large p, small N'' setting. Specifically, when the dimensionality $p$  is larger than the sample size $N$, an intermediate representation is required to avoid (1) numerical stability issues with matrix operations, and (2) the ``curse of dimensionality'' for statistical operations. This intermediate representation can be explicit (as in manifold learning) or implicit (as in kernel machines).  When $N$ is large, even approximation algorithms are required to compute various quantities whose exact solution requires $\mathcal{O}(N^2)$ or $\mathcal{O}(N^3)$ space and/or time.   

Geodesic learning is the process of estimating geodesic distances between pairs of points in a data corpus.  It is crucial (though sometimes implicit) in many state-of-the-art machine learning algorithms.  For example, the first step in many manifold learning algorithms is to estimate the geodesic distance between all pairs of points~\cite{Lee2007-bw}.  
Nonetheless, scant work explicitly addresses geodesic learning, perhaps  because in any real data application the true geodesic distances are not observed, or because a suitable metric for evaluating geodesic learning is not currently available.  Despite this, several disciplines in computer science and machine learning have developed  strategies to partially address the challenges associated with geodesic learning. 

% TODO@VSN what are those challenges?

% \begin{itemize}
% \item 
In computer science, space-partitioning trees are used extensively for quite diverse applications; most relevant to this work are efforts to build trees in support of  efficient geometric queries~\cite{thibault1987set}.  Commonly, space partitioning trees use binary and recursive splits with hyperplanes~\cite{BSP-trees}. These tree structures are usually optimized to learn relative proximities of the observed, noisy measurements, rather than the latent noise-free, and potentially lower dimensional, measurements. 
% \item 
These partition trees from data structures are closely related to decision trees developed in  statistics and machine learning~\cite{Breiman1984-pc}. In fact, extensions to decision trees are established as the de facto standard for classification and regression tasks (even in this age of deep learning), including  random forests~\cite{breiman2001random} and gradient boosting trees~\cite{friedman2001greedy}. These approaches, however, are almost exclusively concerned with supervised, rather than unsupervised learning. 
Decision trees have always been  linked to kernel learning~\cite{Breiman2000-pq}, a realization that recently gained traction in the machine learning literature~\cite{Davies2014-bt, Scornet2016-vf, Balog2016-iv, Shen2018-ca}.
Kernel machines typically assume a kernel, or select one from a finite set or one-dimensional family of kernels, limiting their finite sample performance. 
% \end{itemize}
In manifold learning, many spectral variants start by estimating all pairwise geodesic distances~\cite{scholkopf1997kernel}. These approaches typically  operate on the observed, typically high-dimensional data, and therefore suffer serious drawbacks in the face of additional noise dimensions. 
Although each of the above works is closely related to geodesic learning, to our knowledge none of it explicitly aims to estimate geodesic distances as an end unto itself.  Moreover, those approaches that do estimate geodesic distances typically do not directly evaluate the estimated distances. 

We  propose  \emph{Unsupervised Randomer Forest} (URerF) to achieve near linear space and time complexity, 
%TODO@VSN: should we state the complexity directly? Whenever I see "near-linear" in a paper I get suspicious -- is it O(n log n)? O(n log(log(n)))?, n * inverse Ackermann function?, etc..
%VSN: O(Tdnlog^2n)? YES!
% T trees
% d dimensions
% n log n to sort k features before finding optimal split finding
% finding optimal split: linear in k
% from RerF paper: s Lumberjack’s complexity O(T d log n(n log n+λp)), where λ is the fraction of nonzeros in the p×d random projection matrix. We generally let λ be close to 1/p, giving a complexity of O(T dn log2 n), which is the same as for RF
while approximating the true latent geodesic distances.
Unlike the previously described methods, URerF does not need to compute geodesic distances between all pairs of points. Instead, URerF examines local structure by recursively  clustering data in sparse linear subspaces, building on the recently proposed randomer forest algorithm for supervised learning~\cite{tomita2015randomer}. The randomer forest approach allows URerF to separate meaningful structure in the data from the noise dimensions. We also introduce a spitting criteria, Fast-BIC, which efficiently and exactly computes the Bayesian Information Criterion statistic for an approximate Gaussian mixture model in one dimension.

This manuscript also contributes a  method for evaluating geodesic learning algorithms.  
Most existing manuscripts on manifold learning that explicitly estimate geodesic distances, do not explicitly evaluate the geodesics. Rather, such papers typically embed the data into some low-dimensional space and then  visualize the results.  This approach is  limited in a number of ways: (1) it is  qualitative; (2) when the structure is higher dimensional it may be revealed by the first few dimensions; and (3) it relies on an embedding, which introduces additional computational and statistical complications. 
% TODO@VSN: what complications -- list some?
Sometimes the embedded data are used for subsequent inference tasks, such as classification, which can be quantitatively evaluated.  Such an approach is only able to evaluate performance of a manifold learning algorithm composed with a particular subsequent inferential method, but not the manifold or geodesic learning algorithms directly.  
% TODO@VSN: [I'm not sure I understand what you're trying to say in this sentence. What is the limiting factor, how the algo is composed or that it can only be evaluated indirectly?]
% VSN: i think the above is clear, since there is really only 1 way to compose algo's, but if you think maybe unclear, please fix.

We therefore introduce geodesic precision and recall.  
In contrast to precision and recall as typically defined, geodesic precision and recall quantify the set of nearest neighbors as estimated by the geodesic learning algorithm, with the set of true nearest neighbors on the latent manifolds. 
As a general rule, if a geodesic learning algorithm does poorly on this metric, estimating manifolds from these geodesics has  no hope to perform well on subsequent tasks. Indeed, functions of geodesic precision can provide tight bounds on subsequent classification accuracy \cite{Devroye1997-bd}. 

URerF finds neighbors in the latent low-dimensional space amid additional noise dimensions more effectively than other approaches. Moreover it can do this in a variety of linear and nonlinear settings, with different dimensional submanifolds, and in a real connectome dataset. 

\section{Related Work}

Nonlinear manifold learning approaches, such as Isomap~\cite{tenenbaum2000global}, Laplacian eigenmaps~\cite{belkin2002laplacian} and UMAP~\cite{mcinnes2018umap}, are designed to preserve geodesic distances, and even directly estimate them. Specifically, they follow a three-step process. First, they estimate geodesic distances in the original manifold. This is done by initially constructing a $k$-nearest neighbor or $\epsilon$-neighborhood graph in which the observations (data points) correspond to nodes, and pairwise Euclidean distances between these points correspond to the weights on the edges. 
%TODO@VSN: the first step is to construct a k-nearest neighbor graph or epsilon graph?...and change "distances correspond to edges" to "Euclidean distances determine which pairs of nodes make up edges" or something like that
%VSN: Done
Second, the all-pairs shortest paths of the nodes in the graph are computed. Third, the points are  embedded in a lower dimensional space that ideally preserves these distances. This approach is significantly hampered by the first step, which operates in the original high-dimensional ambient space, since
% This approach has two main shortcomings. 
% First, computing all pairs shortest paths is a computationally intensive algorithm with a runtime complexity of $O(n^3)$ in the number of nodes. Second, for high dimensional datasets, due to the curse of dimensionality, 
Euclidean distances often fail to provide good estimates of distances on the manifold. Moreover, given $n$ datapoints, computing all pairwise distances is $\mathcal{O}(n^2)$ space and time, and all pairwise shortest paths can require $\mathcal{O}(n^3)$, both of which can be cost prohibitive for large sample sizes.

One of the most widely used methods
for nonlinear dimensionality reduction
is   Isomap  \cite{tenenbaum2000global}. 
% , which constructs a low-dimensional embedding of input data by first computing pairwise distances in the observed dimensionality, and then approximating geodesic  distances between data points  in a k-nearest neighbor graph, and finally applying classical multidimensional scaling to the matrix of graph distances.
Isomap is one of the few manifold learning algorithms that has theoretical guarantees for correctly estimating the manifold under certain assumptions~\cite{Silva2003-nl}. 
In the case of many noisy dimensions, however,
Isomap  fails to construct
an accurate nearest-neighbor graphs on the latent manifold. Moreover, Isomap requires storing all point-to-point
graph distances, which incurs space and time complexity quadratic in the sample size.

UMAP is a new algorithm for dimensionality reduction
that efficiently reduces high-dimensional data
to a low dimension using a fuzzy simplicial set representation of the input data points \cite{mcinnes2018umap}.
Like other nearest-neighbor based algorithms,
UMAP first constructs an undirected, weighted k-nearest neighbor 
graph from the input data, then
embeds data points in a low-dimensional space
using a force-directed layout algorithm.
The number of neighbors used to construct the
graph in effect determines the local manifold structure that is to be preserved in the 
low-dimensional layout.
In the force-directed layout approach, 
attractive forces between close vertices
are iteratively balanced with repulsive forces between
vertices that are far apart in the graph
until convergence.
% As we show in our section \ref{sec: exp},
% UMAP is indeed an efficient algorithm but
% is outperformed by our UReRF algorithm when
% the input data contains a high degree of noise.
UMAP builds upon the popular 
t-Distributed Stochastic Neighbor Embedding (t-SNE) algorithm, which  
attempts to preserve original interpoint distances in a 
much lower dimensional space~\cite{Maaten2008-tn}. The Kullback-Leibler divergence
between the distribution of neighbor distances in the higher and lower dimensional spaces is used to determine the optimal mapping of points into the lower-dimensional space.
t-SNE is primarily used to visualize high-dimensional data \cite{maaten2008visualizing},
and cannot be used with non-metric distances. The UMAP algorithm produces similar embeddings to t-SNE in two or three dimensions, 
but scales better in terms of run-time across a wide range of embedding dimensions \cite{mcinnes2018umap}. 

Approximate nearest neighbors algorithms, such as FLANN~\cite{muja2014scalable},  approximate nearest-neighbors in high-dimensional data sets, typically by building binary space-partitioning trees, such as $K$-d trees. These algorithms are designed to estimate the distances in the observed high-dimensional space.  When the true manifold is low-dimensional, and the data are high-dimensional, the additional noise dimensions will be problematic for any of these algorithms. On the other hand, these approaches can achieve near linear space and time complexity. 

This work is inspired by, and closely related to,  random projection trees   for manifold learning~\cite{dasgupta2008random} and vector quantization~\cite{Dasgupta2008-ja}.  The main differences between our approach and theirs is (1) that they use random splits, rather than optimizing the splits; and (2) they use a single tree, whereas URerF uses a forest of many trees.  Nonetheless, their theoretical analysis motivates the geodesic precision metric we establish for quantifying performance of geodesic learning.

Finally, most closely related to our method are existing unsupervised random forest methods, the most popular of which is included in Adele Cutler's RandomForest R package \cite{Shi2006-ka}.  It proceeds by generating a synthetic copy of the data by randomly permuting each feature independently of the others, and then attempts to classify the real versus the synthetic dataset.  As will be seen below, this approach leads to missing surprisingly easy latent structures.

\section{Unsupervised Randomer Forests}
A random forest is an ensemble of decision trees in which each tree is created from bootstrapped samples of the training dataset; that is, each tree is built from a random subset of training data. Each tree $\{h(\textbf{x}, \theta_t)\}, t \in \{1, 2,\ldots, T\}$  has parameters $\theta_t$ that characterize the tree structure, and can be learned from the dataset. Given a set of trees, and a new point $x$, each tree casts a unit vote for its predictions given the  input \textbf{x}. 
%  
% Random Forests can be used for both supervised and unsupervised situations. 
Typically, random forests are used in  supervised machine learning tasks, specifically classification and regression. There have been a few papers reporting on unsupervised random forests  for certain tasks~\cite{shi2006unsupervised}. 
% Our paper differs in the following ways. In particular, for the unsupervised case, 

% estimated as the fraction of the trees in which $x_n$ and $x_j$ appear in the same leaf node. 
Our unsupervised random forest algorithm is based on the original Random Forest algorithm~\cite{breiman2001random} with a few key distinctions. 
First, URerF uses a new splitting criteria, Fast-BIC, that efficiently and exactly computes an approximate Bayesian Information Criterion for a Gaussian Mixture model in one dimension. 
Second, we use the term \textit{randomer} to label our technique, as our splitting methods are  based on random sparse linear combinations of features to strengthen each tree, as originally proposed by Breiman~\cite{breiman2001random}, and later studied further by Tomita et al.~\cite{tomita2015randomer, roflmao}
% in Tomita et al.'s \textit{Randomer Forests} algorithm \cite{tomita2015randomer}. %Additional randomness is therefore incorporated into the forest generation process in candidate dimension generation stage as well as the traditionally randomness-inducing bagging stage. 
% 
Third, we correctly implement a previously proposed method for generating proximity matrices from random forests. In one of the most widely used implementations of Random Forest \cite{LiawRF}, the aggregated normalized proximity matrices of $F$ Random Forests with $T$ trees each is not stochastically equivalent to the aggregated normalized proximity matrices of $T$ Random Forests with $F$ trees each. Our implementation does not suffer from this bug. Furthermore, it is computationally more efficient than previous implementations. %  
% TODO@VSN: prove it with some experiments, for journal paper
These three changes enable  URerF to achieve state-of-the-art performance on both simulated and real data.

\subsection{Overall algorithm} \label{overall}

Given an input data set $x=\{x_1,\ldots, x_N\}$, where $x_n \in \Real^p$, URerF builds $T$ decision trees, each from a random sample of size $m<N$. In each tree, URerF recursively splits a parent node into its two child nodes until some termination specification is met. At each node, URerF generates $d$ 
% very sparse random projection matrix~\cite{Hastie2006-hy}, multiplies the data by it, which results in $d$ 
features to search over.
% , some of which are the original dimensions, others are sparse linear combinations. 
%
Each feature is evaluated based on the splitting criteria described in Section \ref{split}, and the feature with the best score is selected to split the data points into two daughter nodes.  
% The best feature is selected to split on based on the score that results from the splitting criteria described in Section \ref{split}. 
Algorithm \ref{alg:2} describes the procedure used to build unsupervised decision trees (all algorithms are relegated to the appendix).
To evaluate the forest, a proximity matrix is then generated by computing the fraction of the trees in which every pair of elements reside in the same leaf node (Section \ref{prox}).

\subsection{Node-Wise Feature Generation}

% TODO@VSN: write out the way we sample A matrices explicitly, how we use them to transform the data matrix X, etc. refer to breiman's original paper in which he described something similar, and tyler's 2 papers on the subject.
Unlike Breiman's original random forest algorithm, URerF does not choose split points in the original feature space. Instead, we follow the random projection framework of Tomita et al. \cite{tomita2015randomer,roflmao}. 
For $p$-dimensional input data, we sample a $p \times d$ matrix $A$ distributed as $f_A$, where $f_A$ is the projection distribution and $d$ is the dimensionality of the projected space. 
We chose to use the $f_A$ that Tomita et al. empirically found to produce the best performance in the supervised setting: $A$ is generated by randomly sampling from $\lbrace -1, +1 \rbrace$ $\lambda p d$  times, then distributing these values uniformly at random in $A$  \cite{tomita2015randomer}. The $\lambda$ parameter is used to control the sparsity of $A$, and is set to $\frac{1}{20}$, again following the convention of Tomita et al.
Using the randomly sampled $p \times d$ matrix $A$, the data associated with the given node, $X'$, which is a subsample of the original data, is transformed into a $d$-dimensional feature space, where each of the $d$ new features is now a sparse linear combination of the $p$ original features. In other words, each row of $\tilde{X} = A^{T} X'$ represents a projection of the data into a one-dimensional space that is a sparse linear combination of the original feature space. Each of the $d$ rows $\tilde{X}[i,:], i \in \lbrace 1, 2, ..., d \rbrace$ is then inspected for the best split point. The optimal split point and splitting dimension are chosen according to which point/dimension pair minimizes the splitting criteria described in the following section.
 
\subsection{Splitting Criteria} \label{split}

% We compare several splitting criteria to evaluate  unsupervised randomer forests, including: 
% 2-means clustering,
% two-component Gaussian mixture modeling using expectation 
% two-means clustering with the Bayesian Information Criterion (BIC),
% and a soft clustering as defined by the 
% most likely Gaussian Mixture Model (GMM) of two Gaussians with the BIC test.
% The advantage of incorporating the BIC test
% into the splitting mechanism is its ability
% to select the split which results in two maximally distinct clusters. The BIC test outputs a score that measures  how well the datapoints are explained by a Gaussian mixture model with two Gaussians.

\paragraph{Fast, Exact, Univariate, Two-Means Splitting} 
The goal is to find the split that minimizes the sum of the intra-cluster variance on the projected dimension. Typically, k-means problems are solved via Ward's or Hardigan's algorithms~\cite{Ward1963-dm, Hartigan1979-cc}.  Because k-means is NP-hard, in general, these algorithms lack strong theoretical guarantees~\cite{Arthur2007-wn}.
However, in one-dimension, for two-means, there is an exact solution that is much faster and simpler. 
This is available because each decision tree always  operates on one-dimensional marginals.  
First,  sort the data points. 
Then, consider splitting  between all sequential pairs of points;  that is, letting $x_{(s)}$ denote the $s^{th}$ smallest sample,  consider splitting between $x_{(s)}$ and $x_{(s+1)}$ for all $s< N$. The samples to the left of the split point form one cluster, and those to the right form the other cluster.  Estimate the means for each of the clusters using the maximum likelihood estimate (MLE) for all points in that cluster. 
%URerF 
Fast, exact, univariate two-means splitting seeks to find the cutpoint that minimizes the one-dimensional 2-means objective. This splitting criteria was introduced in \cite{dasgupta2008random}.
\begin{align}
\min_{s} \sum_{n=1}^{s} (x_n - \hat{\mu}_1)^2 + \sum_{n=s+1}^{N} (x_n - \hat{\mu}_2)^2.
\end{align}
An immediate limitation of this approach is that it fails to consider feature-wise variance, which can lead to undesirable properties.  For example, if any feature has zero variance, it will always achieve the minimum possible score. 
Although one can rescale each feature independently, doing so can cause problems in unsupervised learning problems when the relative scale of features is important, and the details of how to rescale introduce an undesirable algorithm parameter to tune.

\paragraph{2-Gaussian Mixture Model Splitting with Mclust-BIC}
% TODO@VSN update figures to use the same name as in the text. consider using Fast-BIC and Mclust-BIC
 
In this case, for each feature we fit the data to a two-component Gaussian mixture model (GMM).
An expectation-maximization (EM) is used to jointly estimate all the parameters and latent variables~\cite{Fraley2002-eg}.  
The latent variables, $\{z_{n,j}\}$, denote the probability that sample $n$ is in cluster $j$.  
Letting $N$ be the number of observations and $J$ be the number of Gaussian clusters (in this case, $J=2$), and introducing notation 
$x= \left( x_1, \ldots, x_N\right)$, and
$z=\{z_{1,1}, z_{1,2}, \ldots, z_{N,J}\}$,
the complete likelihood (including the latent variables) is
\begin{equation} %\label{likprob}
    P(x, z;  {\mu},  {\sigma}, {\pi}) = \prod_{n=1}^{N} \prod_{j=1}^{J} \{\pi_j\mathcal{N}(x_n ; \mu_j, \sigma_j^2)\}^{z_{n,j}}.
\end{equation}
% The Gaussian Mixture Modeling (GMM) likelihood equations a re the same as in the above case, except we now relax the binary constraints on the $z_n$.  Specifically, each point has associated with it a non-negative vector $z_n$ that sums to one.  
Each feature is evaluated using the Bayesian Information Criterion (BIC).
BIC is based on the log likelihood of the model given the data, with a regularization term penalizing complex models with many parameters. Concretely, letting $\hat{L}_M$ denote the maximum log likelihood function of a particular model $M$, $\hat{L}_M =p(x;\hat{\theta}_M)$,  where $\hat{\theta}_M$ are the parameters that maximize the likelihood function for model $M$, and $x$ is the observed data.  Letting
$N$ be the sample size (number of data points) and $d_M$ be the number of parameters estimated by the model, then
the BIC score  can be defined as follows:
\begin{equation}
BIC(M) =  - 2\ln(\hat{L}_M)+ \ln(N)d_M.
\end{equation}
The feature  that maximizes the BIC score for a two-component GMM is selected for splitting at each node. The split occurs at the midpoint where the two Gaussians are equally likely. 
Because this approach is standard in the literature, we do not provide pseudocode.
Note that the EM approximates the actual log likelihood, and is only guaranteed to find a local maximum, not the global maximum, rending it sensitive to initialization.  Moreover, the EM algorithm is known to suffer from poor convergence properties in certain settings~\cite{McLachlan2008-sa}.

\paragraph{2-GMM Splitting with Fast-BIC}

Fast-BIC combines the speed of two-means with the model flexibility of Mclust-BIC.
As in two-means, for each feature,  sort all the data, and try all possible splits.  For each split, assign all points below the split to one Gaussian, and all points above the split to the other Gaussian.  Estimate the prior, means, and variances for both clusters using the MLE.  For $j=1$, they are defined by
\begin{align*}
\hat{\mu}_1 = \frac{1}{s}\sum_{n \leq s} x_n, \qquad
\hat{\sigma}_1 = \frac{1}{s}\sum_{n \leq s}||x_{n} - \hat{\mu_j}||^2, \qquad
\hat{\pi}_1 = \frac{s}{N},
\end{align*}
and similarly for $j=2$. 
Under the above assumption,  $z_{n,j}$ is an indicator that data point $x_n$ is in cluster $j$.  In other words, rather than the soft clustering of GMM, Fast-BIC performs a hard clustering, as in two-means.  Thus, if $x_n$ is in cluster $j$, then $z_{n,j}=1$ and $z_{n,j'}=0$ for all $j \neq j'$. 
Given this approximation, the   likelihood can be obtained by summing over the $z$'s
\begin{equation}\label{likprob2}
    P(x;  \mu,  \sigma, \pi) = \sum_{z}\prod_{n=1}^{N} \prod_{j=1}^{J} \{\pi_j\mathcal{N}(x_n ; \mu_j, \sigma_j^2)\}^{z_{n,j}}.
\end{equation}
Noting that  $z_{(n \in (0, s], k = 0)}  = z_{(n \in [s+1, N), k = 1)} = 1$ and $z_{n,j}=0$ otherwise, Equation (\ref{likprob2}) can be simplified to
\begin{align*}
    P(x ; \mu, \sigma, \pi) = %\\
    \prod_{n=1}^{s}\pi_1 \mathcal{N}(x_n ; \hat{\mu_1}, \sigma_1^2) \prod_{n=s+1}^{N}\pi_2 \mathcal{N}(x_n ; \hat{\mu_2}, \sigma_2^2).
\end{align*}
Plugging in the MLE for all the parameters, the  maximum log likelihood function  $\hat{L} =\log P(x ; \hat{\mu}, \hat{\sigma}, \hat{\pi})$ is
\begin{align}\label{loglik1}
   \hat{L} = \sum_{n=1}^{s} [\log \hat{\pi}_1 + \log \mathcal{N}(x_{n} ; \hat{\mu_1}, \hat{\sigma_{1}}^2)] %\\
   + \sum_{n=s+1}^{N} [\log \hat{\pi}_2 + \log \mathcal{N}(x_{n} ; \hat{\mu_2}, \hat{\sigma_{2}}^2)],
\end{align}
Substituting into Equation~\ref{loglik1} and simplifying, we get the following expression for the log likelihood for any given $s$:
\begin{equation}\label{eq3}
  - 2\hat{L}_s =    s\log 2\hat{\pi}_1 \hat{\sigma}_1^2    + (N-s)\log 2\hat{\pi}_2 \hat{\sigma}_2^2 - s\log\hat{\mu_1} - (N-s)\log\hat{\mu_2},
\end{equation} 
in which we have dropped  terms that are not functions of the parameters. 
We further test for the single variance case ($\sigma_1 = \sigma_2$) and use the BIC formula to determine the best case.
Fast-BIC chooses the dimension and split-point that maximizes $\hat{L}_s$.
Pseudocode for this approach is provided in Algorithm \ref{alg:fastbic}.  
Fast-BIC is guaranteed to obtain the global maximum likelihood estimator, whereas the  Mclust-BIC is liable to find only a local maximum. Moreover, Fast-BIC is substantially faster.
This Fast-BIC procedure is, to our knowledge, novel, and of independent interest.

\subsection{Proximity Matrix Construction}\label{prox}

One can build a similarity matrix from any decision tree by asserting that similarity between two  points $x_n$ and $x_j$ is related to some ``tree distance'' between the pair of points in a given tree. Although this is the case for both supervised an unsupervised decision trees, to our knowledge such an approach has not yet been explored for unsupervised trees.  When using a forest, it is natural to average the similarity matrices to obtain the forest's estimate of similarity.  A simple tree distance to use is the $0-1$ loss on whether a pair of points is in the same leaf node. This approach to computing similarities has previously been studied in several supervised random forest papers, connecting random forests to kernel learning~\cite{Breiman2000-pq, Davies2014-bt, Scornet2016-vf, Balog2016-iv, Shen2018-ca}. However, the connection between these similarities and geodesic distances has not yet been established.

More concretely, the proximity matrix $S$ for input data $D \in \mathbb{R}^{n \times d} $ is estimated using the unsupervised random forest by simply counting the fraction of times that a pair of points occurs in the same leaf node in the forest.
Thus, $S(i,j) = S_{ij} = \frac{L_{ij}}{T_{ij}} $,
where $L(i,j)$ is the number of occurrences of points $i$ and $j$ in the
same leaf node,
and $T_{ij}$ is the number of trees in which both point $i$ and point $j$
were included in the bootstrap sample that was used to build the tree. 
We use both the in-bag and out-of-bag samples to estimate the proximity.
% As mentioned above, this approach is consistent with how previous authors computed a random forest induced kernel in the supervised setting~\cite{Breiman2000-pq,Davies2014-bt, Scornet2016-vf, Balog2016-iv, Shen2018-ca}. Since the proximity computation does not depend on the class labels, the same function for computing a kernel can immediately be applied in this unsupervised setting.

%TODO@VSN: consider variations of this for similarity estimation in journal paper: for example, estimate similarity using not only leaf node fraction but some combo of leaf node fraction and shortest path distances within tree 
% VSN: discussed at meeting and these experiments have already been done -- possibly include them in journal version -- ask for location of code for these experiments and results

\subsection{Geodesic Precision and Recall}

% jovo: come back
% The proximity matrices generated by URerF are compared with the Isomap and UMAP distance matrices using geodesic precision and geodesic recall. 

Geodesic precision and recall differ from ``classical'' precision and recall  by virtue of defining the neighbors based on the true latent low-dimensional manifold, rather than the observed (typically higher-dimensional) space.
% Geodesic precision and recall is defined here for two cases: continuous and categorical true neighbors. 
The typical definition of precision and recall are defined relative to a query, the \emph{relevant samples} are those that are ``correct'', where as the \emph{retrieved samples} are those that are returned by the query.  Letting $\cap$ denote set intersection, and $| \cdot |$ denote the cardinality of the set, precision and recall are
\begin{align*}
    \text{precision} &= \frac{| \{ \text{relevant samples}\} \cap \{\text{retrieved samples} \} | }{ | \{ \text{retrieved samples} \} | }, \\
    \text{recall} &=  \frac{| \{ \text{relevant samples}\} \cap \{\text{retrieved samples} \} | }{ | \{ \text{relevant samples} \} | }.
\end{align*}

For geodesic learning, given a data point $x$, a data corpus $\mathcal{D}_N= \{x_1,\ldots, x_N\}$, and a query size $k$, 
the relevant samples  are the $k$ samples from $\mathcal{D}_N$ that are nearest to $x$ based on the true (but unknown) geodesic distance.  
In other words, ``correct'' neighbors is defined by the latent, noise-free, manifold, rather than the observed, typically higher dimensional space.
Given a geodesic learner, the \emph{retrieved samples} are the $k$ samples that the learner reports are nearest.  To compute the geodesic precision and recall for a given learner on a given dataset, average the geodesic precision and recall over each sample point.
Higher precision  and lower recall indicate better estimation of geodesic distances.

We consider two distinct cases: a continuous  geodesic, in which there is a finite geodesic distance between all pairs of points, and a discrete  geodesic, in which there are clusters of points that are not connected at all to other points.  In the latter case (such as a union of spheres), we denote all the points within a given connected component as its neighbors, and all points outside its connected component as not neighbors. 
In the disconnected setting, geodesic precision and recall are identical to one another.

% TODO: all figures should be pdf graphics.
% TODO: for the 4 panel plots, for some reason, the 4 panels are different sizes. would be nice to fix.
\section{Numerical Results}
\label{syntheticdataset}

\subsection{Four  Multivariate Manifold Simulation Settings}

\begin{figure}%[H]
    \centering
        \includegraphics[width=\textwidth]{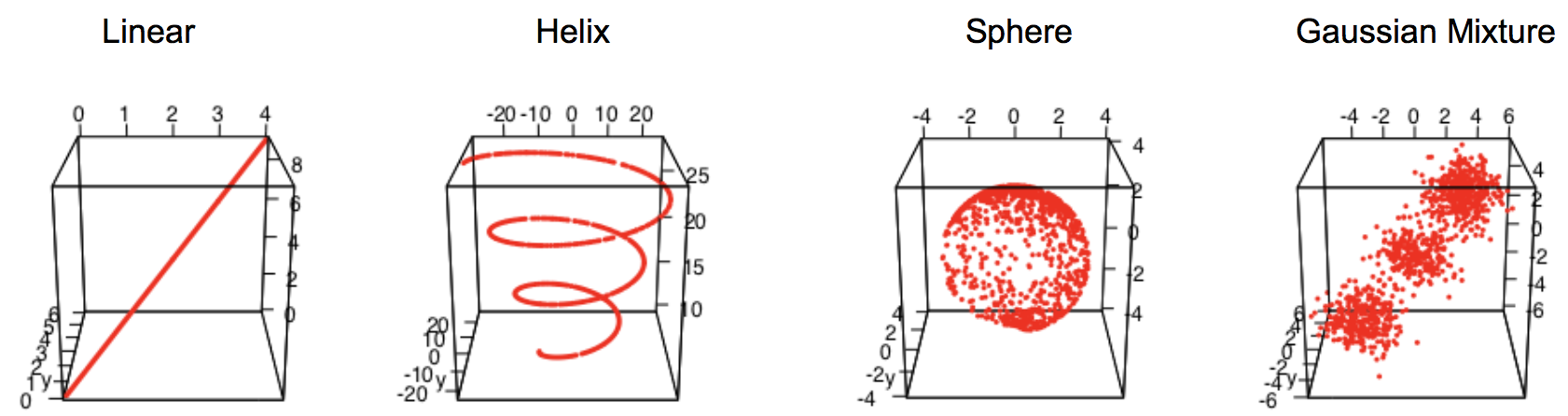} % first figure itself
        \caption{Synthetic datasets for all experiments. In each case, there are 1000 points in 3 signal dimensions, as shown. 
        }\label{fig1}
        % TODO@VSN would be nice to show for more K's, eg, [[2:2:10], [20:20:200]], for this and the next 2 plots.  i wouldn't change the axes.  i would remove the dots.
\end{figure}

We explore geodesic learning using the following four simulations settings, as shown in Figure~\ref{fig1}, each of which span a complementary and interesting case. In the linear case, where learning the geodesic should be relatively easy, Euclidean distance will completely recover the geodesics with no noise. It therefore sets an upper bound on performance. The helix setting is reminiscent of the typical ``swiss jelly roll'' setting popular in manifold learning, but the latent submanifold is one-dimensional embedded into a three-dimensional space.  Here, Euclidean will perform poorly, but various manifold learning algorithms should perform well, as they are designed for this kind of scenario.  The sphere case is interesting because unlike the helix, the true manifold is two-dimensional and could easily be extended to higher dimensions.  Finally, the Gaussian mixture model we suspect will be particularly challenging for the manifold learning algorithms, which typically lack theoretical guarantees for disconnected connected component graphs. Appendix~\ref{app:sims} provides the mathematical details for the four different settings.

% TODO@VSN right here would be a good place to plot estimated geodesics vs true, for each of the 4 settings, for each of the 6 algorithms, for some number of noisy dimensions (like 10).  i'd do 6 rows of 4 columns each.  consider using the hexbin r plot.  there would be a new subsection here about approximating local geodesics. 

\subsection{Choosing the Splitting Criteria and Robustness to Algorithm Parameters}

 \begin{figure}%[H]
    \centering
    % \begin{minipage}{\textwidth}%\label{adele_vs_us}
        % \centering
        \includegraphics[width=\textwidth]{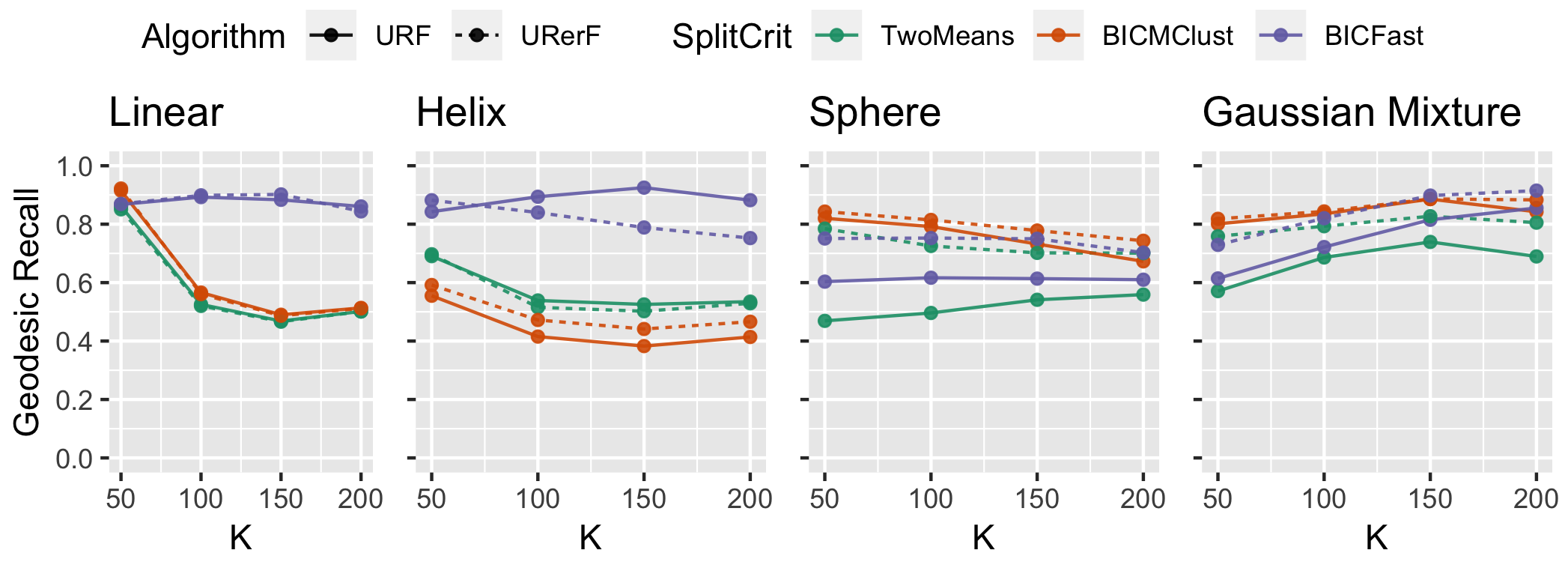} % first figure itself
        \caption{Geodesic recall curves for the three different splitting criteria using both axis-aligned splits (URF; solid lines) and sparse oblique spits (URerF; dashed lines).  In each case there are 1000 points and 3 signal dimensions (no noise dimensions).  In general, URerF with Fast-BIC performs the best or nearly so.  }\label{fig3} 
 \end{figure}

\begin{figure} %[H]
    \centering
    % \begin{minipage}{\textwidth}
        % \centering
        \includegraphics[width=\textwidth]{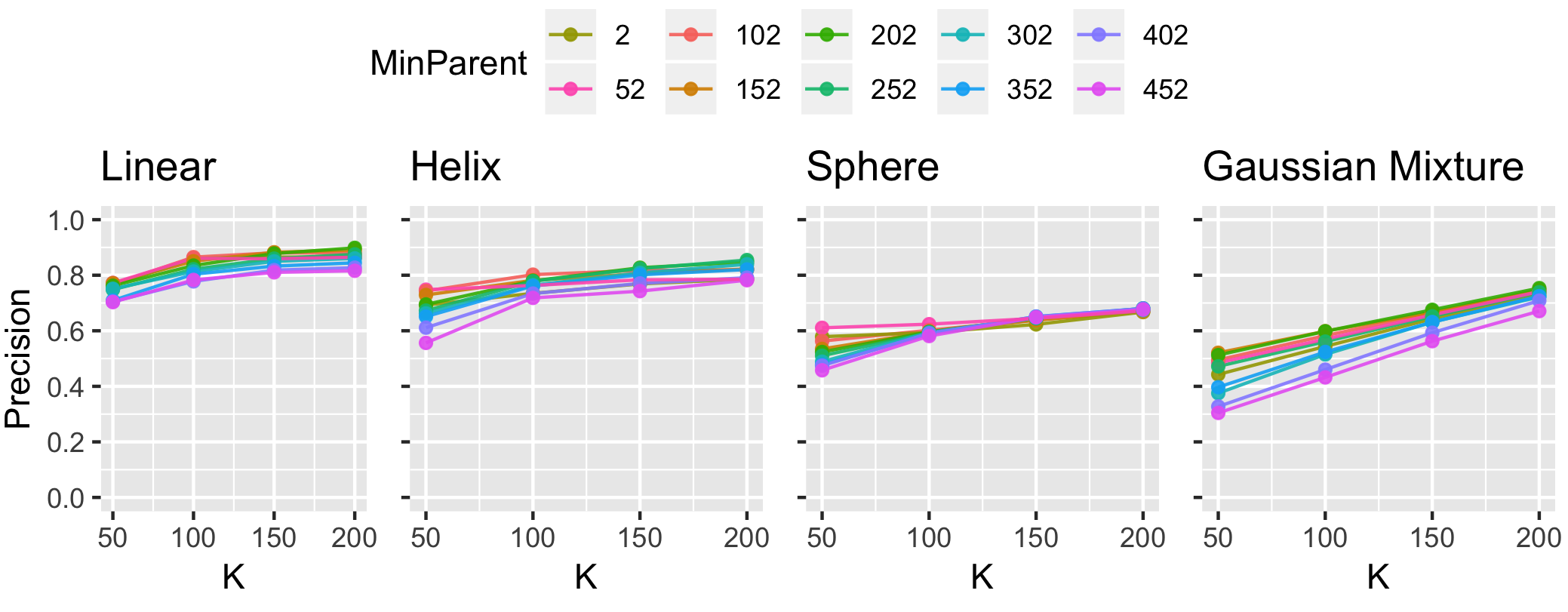} % first figure itself
        \includegraphics[width=\textwidth]{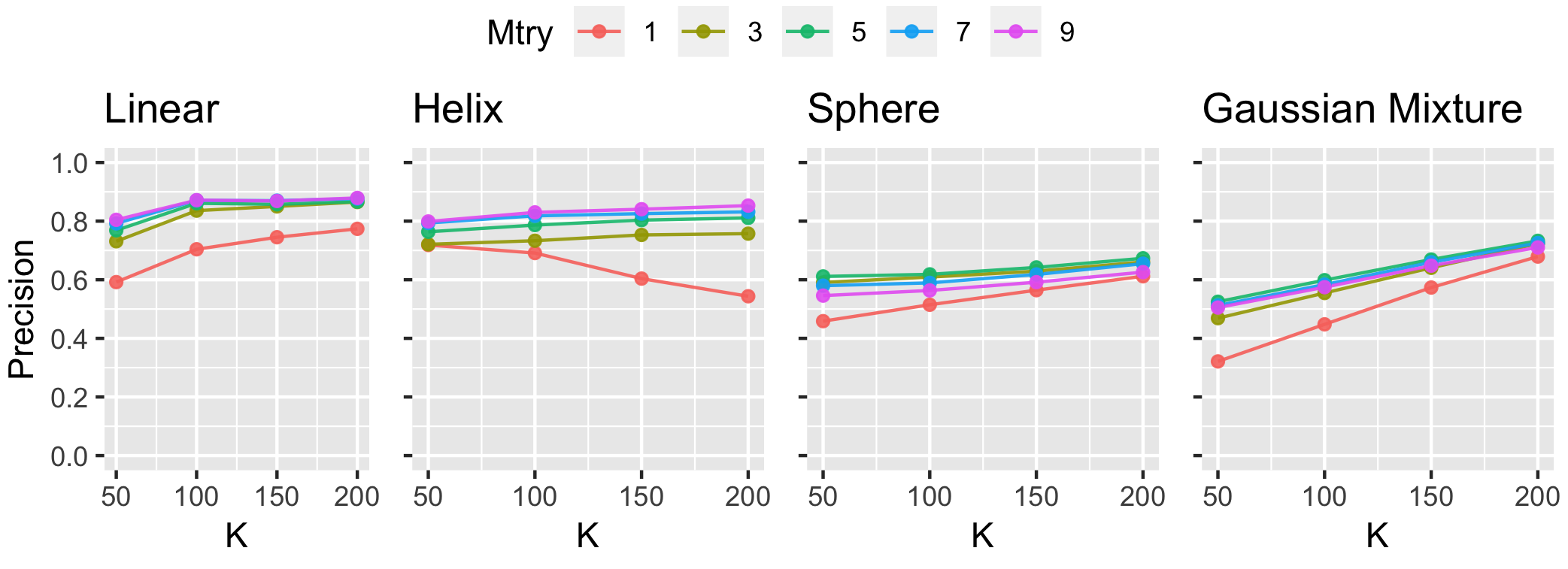} % first figure itself
        \caption{\emph{Top} Geodesic precision versus k for different values of minparent (the smallest splittable node size). Mtry is set to be equal to the square root of the number of features. 
        %, as that was empirically shown to result in a higher precision value. 
        % TODO@VSN: meghana makes the above claim, but i don't know the evidence for it, so i took it out; might be worth exploiring.
        \emph{Bottom} Geodesic precision versus k for different values of mtry (the number of features to test at each node). Minparent was set to be equal to 100. Geodesic precision is robust to large variations in these parameters}\label{fig4}
        % TODO@VSN y-axis should be geodesic precision, not "precision"
        % TODO@VSN would remove titles from the bottom.
    % \end{minipage}\hfill
\end{figure}

For each of the above simulation settings, we sample a thousands points and calculate the geodesic precision using different unsupervised random forest variants.  
Figure \ref{fig3} shows an empirical comparison of the three different splitting criteria described with URF and with URerF. 
In all cases, both BIC approaches (red and blue) outperform two-means splitting criteria (green).  
The solid and dashed lines show the relative performance of URF as compared to URerF  (URerF use sparse oblique splits, that is, splits on linear combinations of the original features), respectively.  In most cases, URerF outperforms URF, as expected based on previous comparisons of sparse oblique splits to axis-aligned splits in supervised random forest~\cite{breiman2001random,tomita2015randomer,roflmao}.
The ramification of these two results is that  in all cases, URerF using Fast-BIC performs as well, or nearly as well, as the other options.  
% 
% In the linear and helix datasets, Fast-BIC performs the best, whereas  in the sphere and gaussian mixture dataset, MClustBIC performs the best. 
% It is expected that MClustBIC outperforms Fast-BIC for settings with somewhat overlapping because its estimates of which point came from which Gaussian are unbiased, whereas Fast-BIC
% This is because the datasets have been generated from multiple distributions (in the case of a mixture of Gaussians) so assigning points to two hard clusters (as in Fast-BIC) will not work as well.
% Because Fast-BIC is as fast as Two-Means, and empirically performs about as well as Mclust-BIC, we use Fast-BIC as our spliiting critera of choice.  
%  Since both have the same computational efficiency, we elect to use URerF for all our experiments. 
Because it performs as well as other options, and runs as fast as two-means, we elect to use URerF+Fast-BIC (hereafter, simply URerF) as our unsupervised decision forest splitting criteria.

In addition to the splitting criteria, each decision tree has  two other important algorithm parameters.  First, minparent, which sets the cardinality of the smallest node that might be split.  Second, mtry, which is the number of features to test at each node. 
Figure \ref{fig4} shows the geodesic precision for different values of minparent and mtry. 
Geodesic precision using URerF  is robust to hyperparameter changes, obviating the need for tuning hyperparameters via a grid search, which can be computationally intensive.  For all future experiments, we set minparent to 100 and mtry to $\sqrt{d}$.

\subsection{URerF is Robust to Noise Dimensions}

\begin{figure}%[H]
    \centering %\offinterlineskip
        \includegraphics[width=\textwidth]{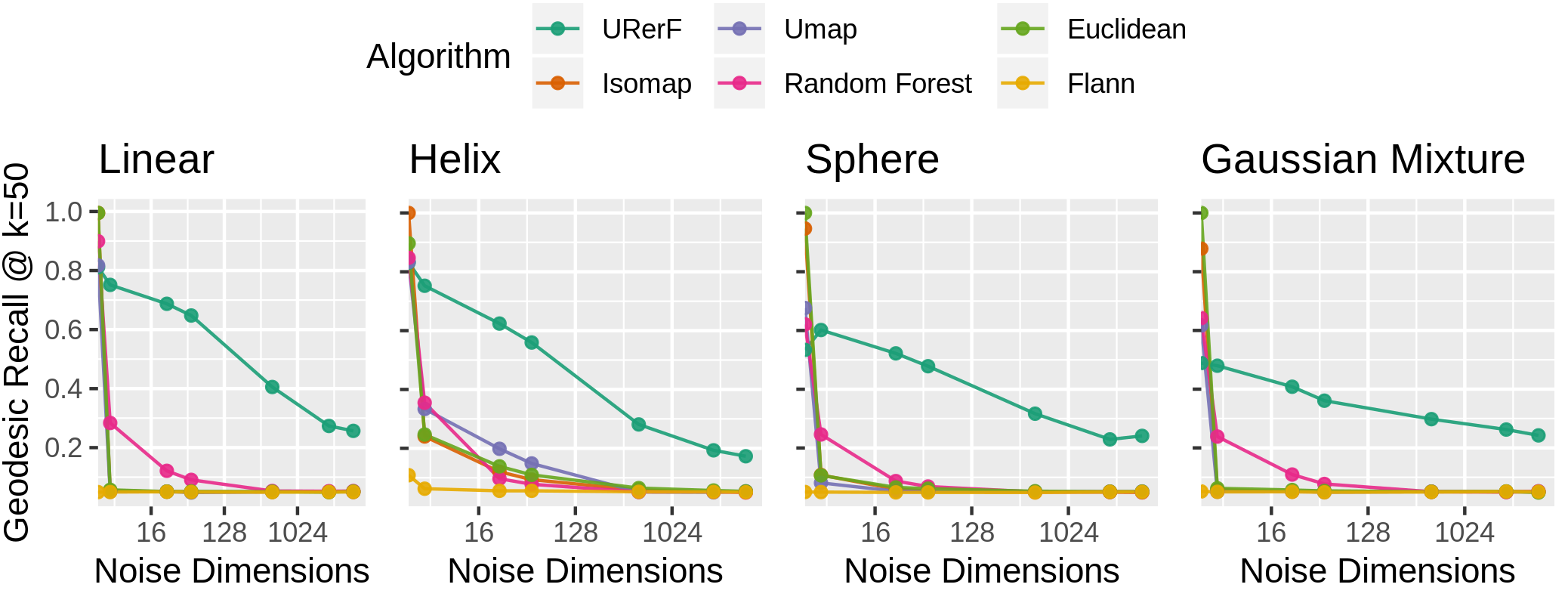} 
        \includegraphics[width=\textwidth]{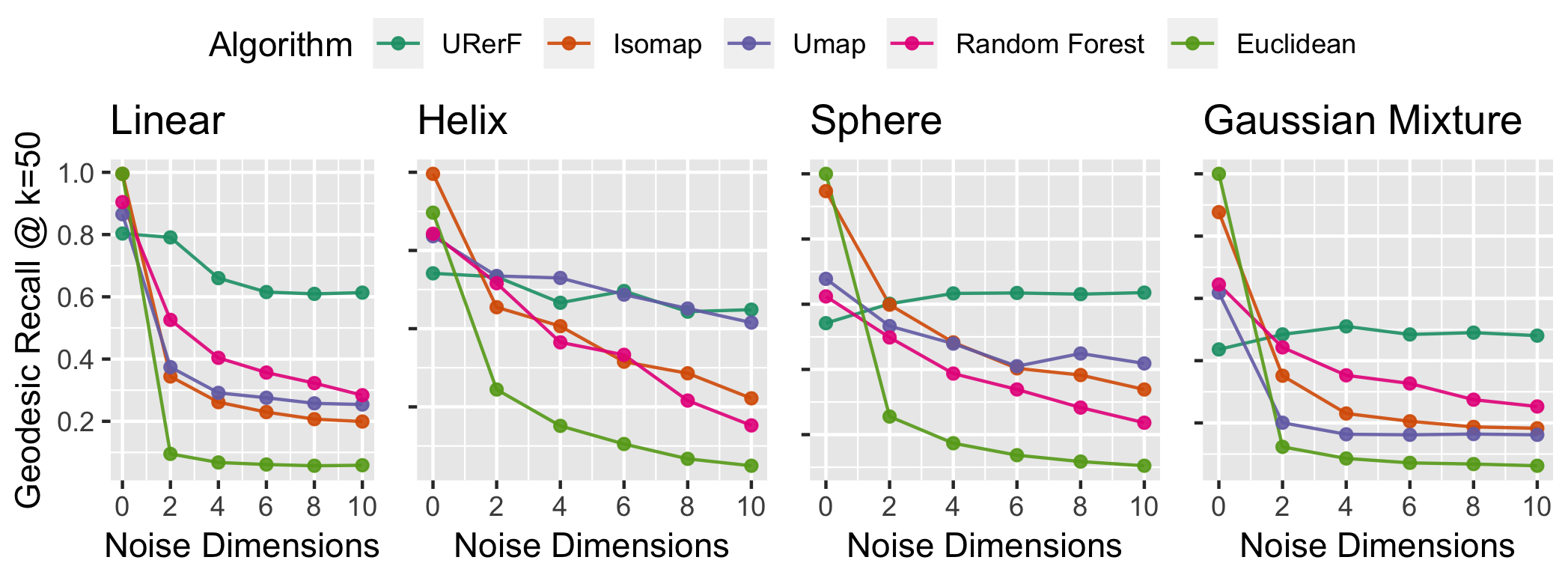} 
        % TODO@VSN remove title and legend for 2nd
        % TODO@VSN add flann to this one.
        \caption{\emph{Top} Geodesic precision at k=50 with varying noise dimension   from 2 to 10,000,  with $N = 1000$ samples.  While all previous state of the art algorithms degrade to chance levels in all settings as the number of noise dimensions increases, URerF never degrades to chance performance for any of the settings. 
        \emph{Bottom} Same as top, but each dimension is linearly rescaled to be between 0 and 1, and x-axis shows a smaller number of dimensions (from 0 to 10).  Although rescaling greatly improves geodesic precision and recall for most algorithms, URerF still achieves much larger geodesic recall for most settings considered. 
        }
        \label{fig2} 
        % TODO remove "noise dimensions" everywhere, on  the far left panels,  replace with "# of noise dimensions" 
        % TODO the x-axis for the top plots should start at 1 and end at 10,000, if that is the upper bound.  can also have a marker in there for 10, 1000.
\end{figure}

% TODO@VSN: say that results are robust to different reasonable choices for k. are they though? experiment for journal?

To see that URerF is robust to high dimensional noise, Gaussian noise with varying dimensions $d'$ are concatenated onto the simulated datasets. Specifically, for each data point $x_n \in \mathbb{R}^d$, generated noise  $y_n \overset{iid}{\sim} \mathcal{N}(0, c \mathbb{I})$ where $y_n \in \mathbb{R}^{d'}$ is concatenated onto $x_n$,   ($c=70$ in the following experiments),
 and $\mathbb{I}$ is the $d' \times d'$ identity matrix.  The new data points with noise are thus: $\tilde{x}_n = [x_n^{\top} | y_n^{\top}]^{\top} \in \mathbb{R}^{d+d'}$. 
Each algorithm's proximity matrices are  computed on the $\tilde{x}$'s,  and compared with geodesic distance matrices to obtain geodesic precision and recall.

Figure \ref{fig2} shows the geodesic recall @ k=50 as a function of the number noise dimensions
for  Isomap, UMAP, random forests, Euclidean distance, FLANN, and URerF. URerF performs well even with the addition of high dimensional noise dimensions. 
% TODO@VSN: is there a better way to say the above? gabi thought so.
The other state-of-the-art algorithms achieve a higher geodesic recall than URerF in the absence of noise dimensions,  but  degrade much more quickly than URerF upon the addition of noise dimensions. FLANN and Euclidean distance degrade the fastest, followed by Isomap and UMAP. This suggests that typical approximate nearest neighbor algorithms  (which are approximating the distance in the ambient space) will perform poorly on recalling the desired items when the data live near a low-dimensional manifold.  The top panel shows the geodesic recall curves for adding up to 10,000 dimensions.  The performance of all the  algorithms except URerF degrades to chance levels in all four settings, whereas URerF maintains a geodesic recall far above chance levels. 
The bottom panel shows geodesic recall after normalizing each of the dimensions (i.e., linearly rescaling each feature to be between 0 and 1). 
Other algorithms are very sensitive to dimension rescaling, and performance may  improve  as a result. 
However, URerF consistently performs better, even without rescaling, as long as a few noise dimensions are added.

% TODO: for the plots with 5-7 colors, the 2 greens are too close for comfort, i'd rather use the qualitative color palette described here: https://bitsandbrains.io/2018/09/08/figures.html, including RED for URerF

\subsection{URerF Estimates Geodesic Recall on Drosophila Connectome}

\begin{figure}
  \centering
%   \begin{minipage}[b]{0.6\textwidth}
    \includegraphics[width=0.47\textwidth]{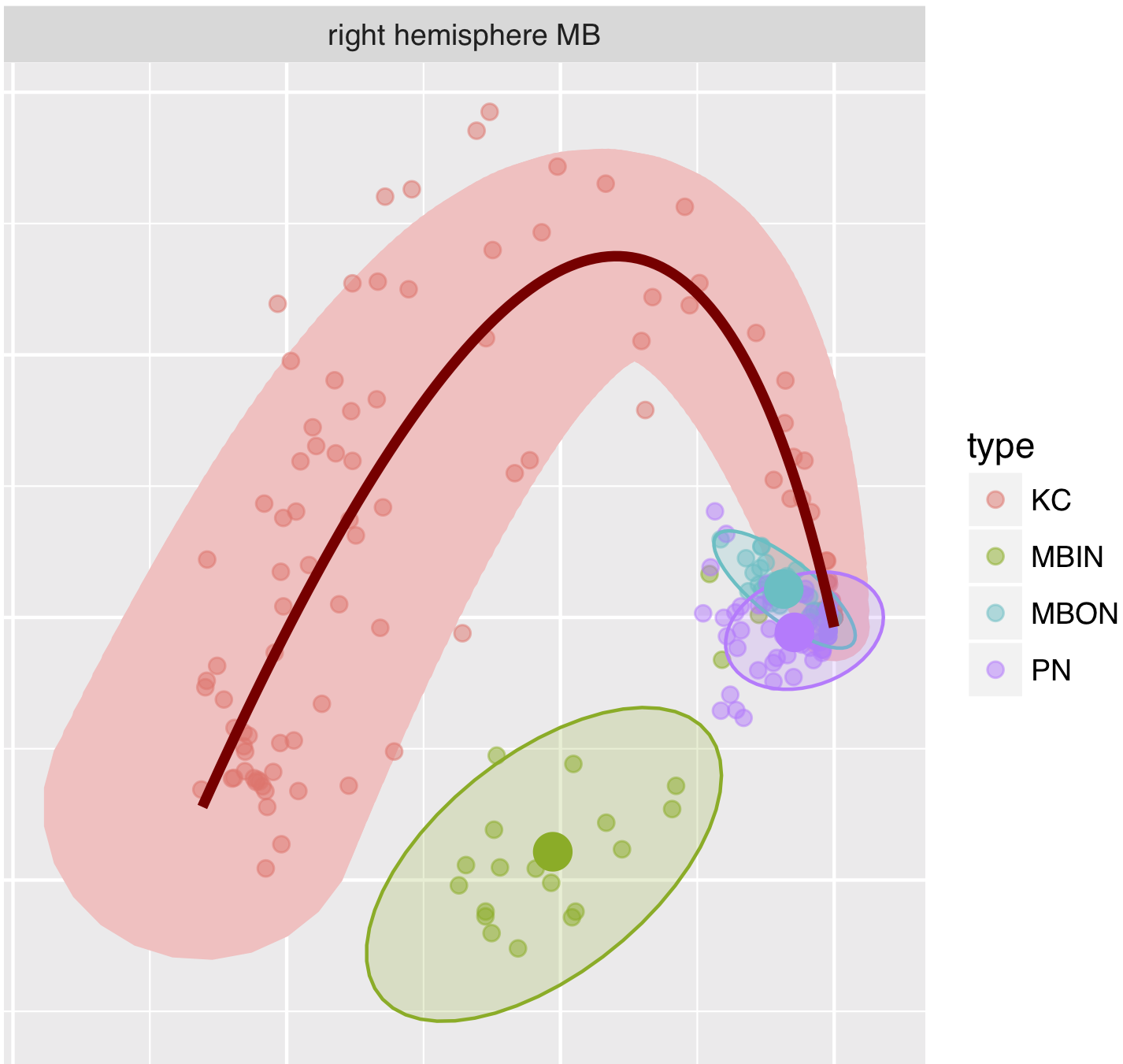}
    % \caption{Drosophila}
%   \end{minipage}
%   \hfill
%   \begin{minipage}[b]{0.4\textwidth}
    \includegraphics[width=0.47\textwidth]{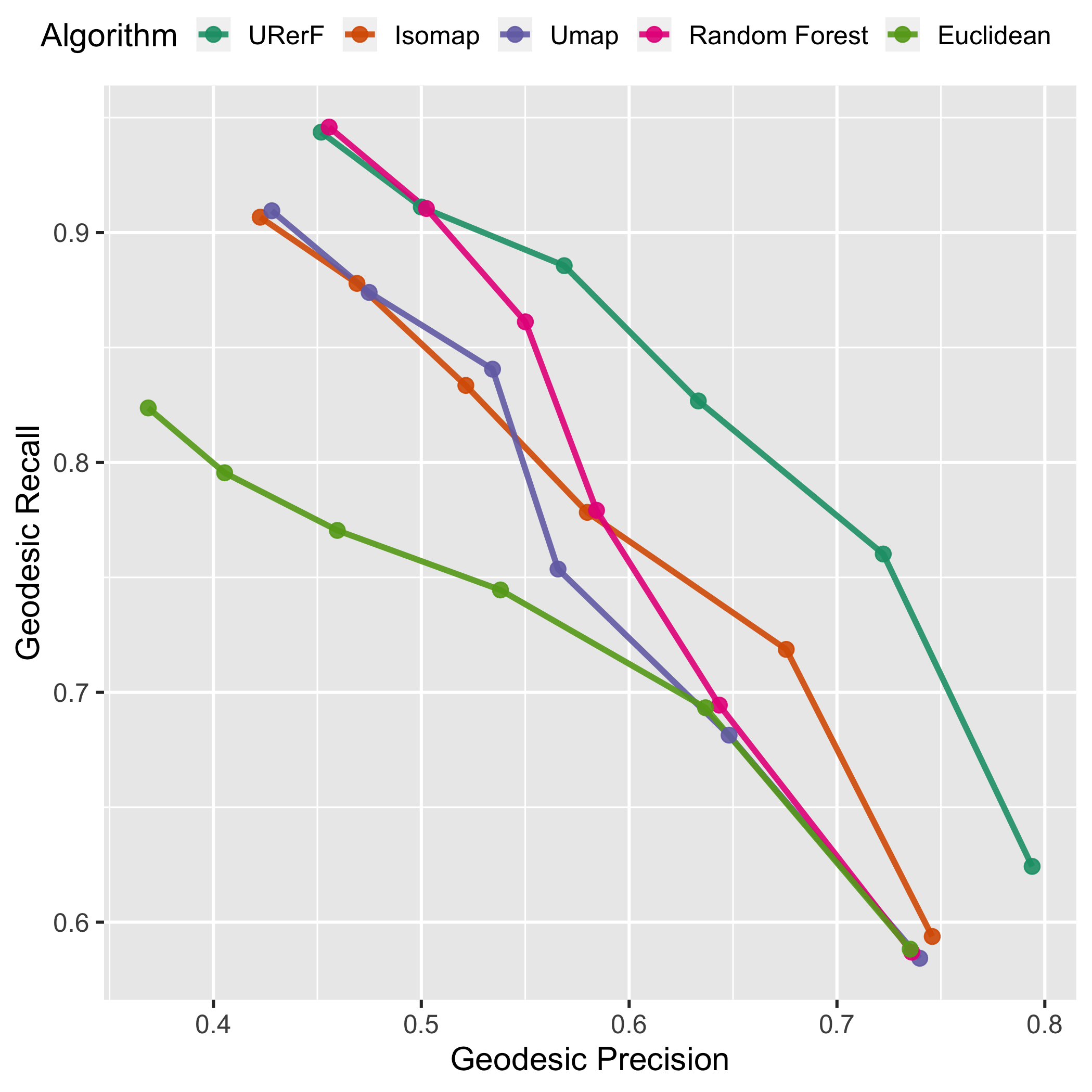}
    \caption{\emph{Left} The right Drosophila connectome after adjacency spectral embedding into six-dimensional space, just showing two of the dimensions. \emph{Right}  Geodesic precision versus geodesic recall for various algorithms using cell type as the true label. URerF achieves a higher recall for essentially all precisions. The values of k for this experiment range from 50 to 250 with increments of 50} \label{fig5}%[H]
%   \end{minipage}
    % TODO@meghana which values of k did we report. i think 50, 100, 150, 200, 250  
    %Jovo you are right.
    % TODO@VSN let's report on k=1,2,...,10, 20, ...,100, 150, 200, 250.
    % TODO@VSN fonts are too small.
\end{figure}

The study of brain networks, or connectomics, is quickly emerging as an important source of real world data challenges~\cite{vogelstein2019connectal}.  Recently, the entire larval Drosophila mushroom body connectome--the learning and memory system of the fly---was estimated and released~\cite{eichler2017complete}.  
It was obtained via manual labeling and semi-automatic machine vision segmentation of serial section transmission electron microscopy. 
The 200 nodes of this connectome correspond to 200 distinct neurons in the mushroom body. 
There are roughly 75000 edges, defined as present between a pair of neurons whenever there exists as least one synapse between them. The edges connect vertices in four known classes of cells: kenyon cells, input neurons, output neurons, and projection neurons~\cite{priebe2017semiparametric}.   
% This connectome contains the entirety of  intrinsic neurons called Kenyon cells and all of their pre- and post-synaptic partners. 
% An important real world data set that we performed experiments on is the connectome of an 
A semiparametric analysis of the connectome, using adjacency spectral embedding~\cite{sussman2012consistent}, results in a six-dimensional latent representation of each node~\cite{priebe2017semiparametric}.  Because this connectome is directed, the first three dimensions correspond to ``outgoing'' latent features, whereas the next three correspond to ``incoming'' latent features. Figure \ref{fig5} (left) shows two of the six dimensions.
Figure \ref{fig5} on the right shows the geodesic precision versus geodesic recall for various algorithms using cell type as the true label.
URerF achieves a higher recall at essentially all precision levels.

% Following the semiparametric spectral modelling of the larval Drosophila mushroom body connectome, it is observed that it can be represented as a directed graph on four neuron types. We observe that 
% URerF is able to accurately compute the nearest neighbor based on similar neurons. 
% right Drosophila mushroom body connectome after adjacency spectral embedding into 6 dimensional space, just showing two of the dimensions. 

\section{Discussion}

We proposed a geodesic distance learning method using Unsupervised Randomer Forests (URerF), as well as a  splitting rule called Fast-BIC. URerF  is empircally robust to noise dimensions, as demonstrated by several different simulation settings, many different added noise dimensions, and  the real-world \emph{Drosophila} connectome. While here we address geodesic learning explicitly, 
geodesic learning is an essential statistical primitive for many subsequent inference tasks.  For example, manifold learning, high-dimensional clustering, anomaly detection, and vertex nomination~\cite{yoder2018vertex} all rely on geodesic learning. More generally, any \emph{ranking} problem is essentially a geodesic learning problem. 
%TODO@VSN: How so? Elaborate how geodesic learning is like ranking.
Moreover, while we only considered unsupervised geodesic learning, the ideas presented here immediately lend themselves to supervised geodesic learning as well.  

We did not explore any theoretical claims associated with the algorithms presented here.  Indeed, we did not even evaluate whether any of these algorithms approximate the precise geodesic. Rather, our metric is concerned purely with getting the geodesic neighbors correct.  
However, prior work using random projections to learn low-dimensional manifolds~\cite{dasgupta2008random}, and for vector quantization~\cite{Dasgupta2008-ja}, have certain theoretical guarantees associated with the intrinsic dimension of an assumed latent manifold. It seems that those guarantees could relatively easily be transferred to this setting.  Another potential direction  would be to address the theoretical bounds that geodesic learning can provide with respect to Bayes optimal performance, on both unsupervised and supervised learning problems.  For example, 1-nearest neighbor provides tight bounds on Bayes optimal classification~\cite{Devroye1997-bd, biau2008consistency}.  Ideas presented in previous work on bounding Bayes performance using ranking algorithms could also extend to this setting.

%TODO@VSN: Either say something about or experiment with noise in the latent dimensions, versus perfect data in latent dimensions plus added noise dimensions, for journal

URerF is  available as part of the open source package ``RerF'' which includes a Python as well as an R package, available at \url{https://neurodata.io/rerf}.

\section*{Acknowledgements}

The authors are grateful for the support by the D3M program of the Defense Advanced Research Projects Agency (DARPA), and  DARPA's Lifelong Learning Machines program through contract FA8650-18-2-7834.

\bibliographystyle{ieeetr}
\bibliography{DEML-bib}

\newpage
\appendix
\section{Algorithms} \label{sec:algorithms}

% TODO@VSN i like pseudocode to look like this, please update for journal
% 
% \begin{algorithm}
% \caption{Permutation Test.
% This algorithm uses the random permutation test with $r$ random permutations for the p-value,
% requiring $\mc{O}(rn^2 \log n)$ for \Mgc. In real data experiments we always set $r=10$,$000$. Note that the p-value computation for any other global generalized correlation coefficient follows from the same algorithm by replacing \Mgc~with the respective test statistic.
% }
% \label{alg:pval}
% \begin{algorithmic}[1]
% \Require A pair of distance matrices $(A, B) \in \Real^{n \times n} \times \Real^{n \times n}$, the number of permutations $r$,
% and \Mgc~statistic $\GG^*$ for the observed data.
% \Ensure The p-value $pval \in [0,1]$.
% \Function{PermutationTest}{$A$, $B$, $r$, $\GG^*$}
% \For{$t:=1,\ldots,r$}
% \State $\pi=\textsc{RandPerm}(n)$ \Comment{generate a random permutation of size $n$}
% \State $\GG^{*}_{0}[t]=\textsc{MGCSampleStat}(A, B(\pi,\pi))$ \Comment{calculate the permuted \Mgc~statistic}
% \EndFor

% \State $pval(\GG^*) \rto \frac{1}{t}\sum_{t=1}^{r}\mb{I}(\GG^{*} \leq \GG^{*}_{0}[t])$  \Comment{compute p-value of \Mgc}

% \EndFunction
% \end{algorithmic}
% \end{algorithm}

%%%%%%%%%%%%%%%%%%%%%%%%%%%%%%%%%%%%%%%%%%%algo 2
\begin{algorithm} [H]
\caption{Build an unsupervised random decision tree. Using sparse linear combinations of features for 1-dimensional projections, find the best splitting point among $d$ of such projections. }
\label{alg:2}
\begin{algorithmic}[1]
\Procedure{BuildTree}{$X, d, \Theta$}
\\ \noindent \textbf{Input: } \\
$X$ : a subset the of training data of dimension $p$\\
$d$ : dimensionality of the projected space \\ 
$\Theta$: set of split eligibility criteria

\noindent \textbf{Output: } a tree \textbf{t} \\

\If{$\Theta$ not satisfied}
	\State \textbf{return} LeafNode($X$) \Comment{Create a leaf node}
\Else
	\State $A \gets \left[a_1 ... a_p \right] \sim f_A$ \Comment{sample random 
	$p \times d$ matrix}
	\State $\tilde{X} = A^T X$ \Comment{random projection into new feature space}
	\State min\_t* $ \leftarrow \infty$ 
	\For{  $ i \in \lbrace 1,...,d\rbrace $} \Comment{test each projected dimension for optimal split}
		\State $\tilde{X}^{(i)} \leftarrow  \tilde{X} \left[ :, i  \right] $
		\State (midpt, t*) = \textbf{ChooseSplit}($\tilde{X}^{(i)}$)  \Comment{Use either Algorithm~\ref{alg:twomeans} or \ref{alg:fastbic}}
		\If{ (t* $< $min\_t*)} \Comment{store best splitting dimension and split point}
			\State bestDim = i
			\State splitPoint = midpt
		\EndIf
	\EndFor
		
	\State $X_{\textnormal{left}} = \lbrace x \in X |  x( \textnormal{bestDim} )  <  \textnormal{splitPoint} \rbrace$
	\State $X_{\textnormal{right}} = \lbrace x \in X |   x(\textnormal{bestDim})  \geq \textnormal{splitPoint} \rbrace$

	\State Daughters.Left = BuildTree($X_{\textnormal{left}},  d, \Theta$)
	\State Daughters.Right = BuildTree($X_{\textnormal{right}},  d, \Theta$)
	 
	\State \textbf{return} Daughters

\EndIf
\EndProcedure
\end{algorithmic}
\end{algorithm}

%%%%%%%%%%%%%%%%%%%%%%%%%%%%%%%%%%%%%%%%%%%%%%%%algo3
\begin{algorithm} [H]
% \label{alg:3}
\caption{Find the optimal split of one-dimensional data, in terms of the two-means objective.}\label{alg:twomeans}
\begin{algorithmic}[1]

\Procedure{TwoMeans1D}{$Z$}\\
\noindent\textbf{Input: } $Z \in \Real^n$: set of one-dimensional input values, one per $n$ data points \\
\noindent\textbf{Output: } Optimal two-means split point, splitPoint, between points in $Z$ and the corresponding sum of squared distances to the two means, minVars 
\State  $\hat{\mu_1} \gets \min(Z) $ 
\State $\mathcal{C}_1 \gets \lbrace \hat{\mu}_1 \rbrace $
\State $\mathcal{C}_2 \gets Z \setminus$ $\mathcal{C}_1$ 
\State $\hat{\mu_2} \gets\frac{1}{n_2} \sum_{ z_i \in \mathcal{C}_2 } z_i $  \Comment{mean of $\mathcal{C}_2$}
\State vars $\gets \sum_{j=1}^2 \sum_{z_i \in C_j} (z_i - \mu_j)^2$ \Comment{sum of variances}
\State minVars $\gets$ vars

\While{ $\mathcal{C}_2 \neq \emptyset$ }
	\State $z \gets \min (\mathcal{C}_2)$
	\State $\mathcal{C}_1 \gets \mathcal{C}_1 \cup \lbrace z \rbrace$
	\State $\mathcal{C}_2 \gets \mathcal{C}_2 \setminus \lbrace z \rbrace$ 
	\State $\hat{\mu_1} \gets \frac{1}{n_1} \sum_{ z_i \in \mathcal{C}_1 } z_i $  \Comment{mean of $\mathcal{C}_1$}
	\State $\hat{\mu_2} \gets\frac{1}{n_2} \sum_{ z_i \in \mathcal{C}_2 } z_i $  \Comment{mean of $\mathcal{C}_2$}	
    %\State midpt $ \gets ( \hat{\mu_1} + \hat{\mu_2} ) / 2$
	\State vars $\gets \sum_{j=1}^2 \sum_{z_i \in \mathcal{C}_j} (z_i - \mu_j)^2$ 	
	\If{ vars $<$ minVars }
		\State minVars $\gets$ vars
		\State splitPoint $\gets (\max(\mathcal{C}_1)+\min(\mathcal{C}_2))/2 $ \Comment{Midpoint between $\mathcal{C}_1$ and $\mathcal{C}_2$} % ( \hat{\mu_1} + \hat{\mu_2} ) / 2$
	\EndIf	
   \EndWhile
\State \textbf{return} (splitPoint, minVars)

\EndProcedure
\end{algorithmic}
\end{algorithm}

%%%%%%%%%%%%%%%%%%%%%%%%%%%%%%%%%%%%%%%%%%%%algo4
\begin{algorithm}[H]
\caption{Fast-BIC1D: Find the optimal split, in terms of BIC score, of one-dimensional data with two classes.}
\label{alg:fastbic}
\begin{algorithmic}[1]

\Procedure{Fast-BIC1D}{$Z$}\\ 
\textbf{Input: } $Z \in \Real^n$: data matrix containing $n$ points in $1$ dimension \\
\textbf{Output: } (splitPoint, minBIC): The midpoint and estimated BIC score that correspond to the best partition between the points in $Z$
\State  $\hat{\mu}_1 \gets \min(Z) $ 
\State $\mathcal{C}_1 \gets \lbrace \hat{\mu_1} \rbrace $ 
\State $\mathcal{C}_2 \gets Z \setminus$ $\mathcal{C}_1$ 
\State $\hat{\mu}_2 \gets\frac{1}{|C_2|} \sum_{ z_i \in \mathcal{C}_2 } z_i $  \Comment{mean of $C_2$}
\State BIC\_curr $\gets \sum_{j=1}^2 \sum_{z_n \in \mathcal{C}_j} (z_n - \mu_j)^2$ 
\State minBIC\_curr $\gets$ dists\_sq

\While{ $\mathcal{C}_2 \neq \emptyset$ }
	\State $z \gets \min (\mathcal{C}_2)$
	\State $\mathcal{C}_1 \gets \mathcal{C}_1 \cup \lbrace z \rbrace$
	\State $\mathcal{C}_2 \gets \mathcal{C}_2 \setminus \lbrace z \rbrace$ 
	\For{j = 1, 2}
	\State $n_j = |\mathcal{C}_j|$
	\State $\hat{w}_j = n_j / n$
	\State $\hat{\mu_j} \gets \frac{1}{n_j} \sum_{ z_i \in \mathcal{C}_j } z_i $  \Comment{mean of $\mathcal{C}_j$} 
% 	\State $\hat{\mu_2} \gets\frac{1}{n_2} \sum_{ z_i \in \mathcal{C}_2 } z_i $  \Comment{mean of $\mathcal{C}_2$}	
    \State $\hat{\sigma_j}^2 \gets \frac{1}{n_j} \sum_{z_i \in \mathcal{C}_j}(z_i - \hat{\mu_j})^2 $  \Comment{variance of $\mathcal{C}_j$}
    % \State $\hat{\sigma_2}^2 \gets \frac{1}{n_2}\sum_{z_i \in \mathcal{C}_2}(z_i - \hat{\mu_2})^2 $
    \EndFor
    
	\State $\hat{\sigma}_\textrm{comb}^2 \gets \frac{1}{n}\sum_{j=1}^2 \sum_{z_i \in \mathcal{C}_j} (z_i - \hat{\mu}_j)^2$ 	
	
	\State BIC\_diff\_var  $\gets -2(n_1\log \hat{w}_1 - \frac{n_1}{2}\log 2\pi \hat{\sigma}_1^2 - n_2\log \hat{w}_2 + \frac{n_2}{2}\log 2\pi \hat{\sigma_2}^2)$
% 	+ n \ln(3) $
	
	\State BIC\_same\_var $\gets -2(n_1\log \hat{w}_1-\frac{n_1}{2}\log 2\pi \hat{\sigma}_\textrm{comb}^2 - n_2\log \hat{w}_2 + \frac{n_2}{2}\log 2\pi \hat{\sigma}_\textrm{comb}^2)$ 
% 	+ n \ln(2)$
	% TODO@meghana: i don't understand the log(3) vs log(2) scaling, so i removed them.  also, the optimal split does not seem to depend on them.
	%Jovo , yes they are just scaling . You are right the optimal split doesnt depend on them
	\State BIC\_curr $\gets \min$ (BIC\_same\_var,  BIC\_diff\_var)
	\If{ BIC\_curr $<$ minBIC}
		\State minBIC $\gets$ BIC\_curr
		\State splitPoint $\gets (\max(\mathcal{C}_1)+\min(\mathcal{C}_2))/2 $ \Comment{Midpoint between $\mathcal{C}_1$ and $\mathcal{C}_2$} % ( \hat{\mu_1} + \hat{\mu_2} ) / 2$
	\EndIf	
   \EndWhile
\State \textbf{return} (splitPoint, minBIC)
\EndProcedure
\end{algorithmic}
\end{algorithm}

 \section{Simulation Settings} \label{app:sims}
 
 \begin{itemize}
    \item \textbf{Linear}:  each point $x$ is parameterized by $p=(4t, 6t, 9t)$, with $t \in (0, 1)$ where $t$ is sampled from a grid with equal spacing.  
    \item \textbf{Helix}:  each point $x$ is parameterized by $p=(t\cos(t), t\sin(t), t)$, with $t \in (2\pi, 9\pi)$ on an equally spaced grid. 
    \item \textbf{Sphere}:  each point $x$ is parameterized by $p=(r\cos(u)\sin(v), r\sin(u)\sin(v), r\cos(v))$, with $u \in (0, 2\pi)$, $v \in (0, \pi)$ and $r=9$ where $u$, $v$ are sampled form a grid with equal spacing.     
    \item \textbf{Gaussian Mixture}:  each point $x$ is drawn from a mixture of Gaussian distributions: $\sum\limits_{j=1}^{3} \hat{w}_j \mathcal{N}(\mu_j, \Sigma),$ with $(\pi_1, \pi_2, \pi_3) =(0.3, 0.3, 0.4)$, 
    $(\hat{\mu_1}, \hat{\mu_2}, \mu_3) = ( \begin{bmatrix} -3 \\ -3 \\-3\end{bmatrix}, \begin{bmatrix} 0 \\ 0 \\0\end{bmatrix}, \begin{bmatrix} 3 \\ 3 \\3\end{bmatrix})$ and 
    $\Sigma = \mathbb{I}$ is the identity matrix. 
    % \begin{pmatrix} 1 & 0 & 0\\ 0 & 1 & 0\\ 0 & 0 & 1 \end{pmatrix}    $.
\end{itemize}

\section{Fast-BIC1D Derivation}

The BIC test is an alternative to the two means split criteria. We rank potential splits on the BIC score obtained assuming a model with a mixture of two Gaussians; the elements along a given dimension to the left of the cut point belong to one Gaussian and the elements to the right belong to another Gaussian (the elements are in sorted order). For example, if the elements are [1, 3, 4, 6] along a dimension, possible splits are [[1],[3, 4, 6]] , [[1, 3],[4, 6]] , [[1, 3, 4],[6]]. In the case [[1, 3],[4, 6]], we assume the model to comprise of two Gaussians, where [1,3] are sampled from the first and [4,6] are sampled from the second. We choose the split that results in the lowest BIC score. We then repeat the process for all dimensions and choose the dimension that gives the split resulting in the lowest BIC score. We use the following notation in the below derivation:

% \subsubsection{Notation}
\begin{itemize}
    % \item subscript 1 refers to the gaussian model of which the elements to the left of the cut point are sampled (cluster 1)
    % \item subscript 2 refers to the Gaussian model of which the elements to the left (right?) of the cut point are sampled (cluster 2).
    \item $Z$ : is the set of $n$ points in one-dimension
    \item $s$ : is the current split point
    \item $n_j$ : Number of elements in cluster $j$
    % \item $n_2$ : Number of elements in cluster 2
    \item $w_j$ : probability of choosing cluster $j$,  $w_j=\frac{n_j}{n}$
    % \item $\pi_2$ : probability of choosing cluster 2 = $\frac{N_2}{N_1+N_2}$
    \item $\mu_j$ : mean of cluster $j$
    % \item $\mu_2$ : mean of cluster 2
    \item $\sigma_{1}^j$ : variance of cluster $j$
    % \item $\sigma_{2}^2$ : variance of cluster 2
\end{itemize}

% \subsubsection{Computing the log likelihood}

Computing the log likelihood is slightly different here than for the regular GMM. This is because we already know which element belongs in which cluster, so we can compute the log likelihood directly, without the  EM step. 

\begin{align}
    \log \ell(Z) &= \log \ell (\{ z_n\}_{n < s}) + \log \ell (\{z_n\}_{n > s}) \nonumber \\
% \end{align}
% \begin{equation}
    \log \prod_n P(x_n;\mu, \sigma^2, w) &= \sum_{n < s} [\log w_1 + \log \mathcal{N}(x_{n} ; \mu_1, \sigma_{1}^2I)] + \sum_{n> s} [\log w_2 + \log \mathcal{N}(x_{n} ; \mu_1, \sigma_{2}^2I)]
\end{align}

Substituting the expression for $w_1$ and $w_2$ and expanding the $\log$ Gaussian, we get the following:
\begin{multline}\label{eq2}
    \log \prod_n P(x_n; \mu, \sigma^2, w) =\\  n_1\log w_1 - \frac{n_1}{2}\log 2 \pi \sigma_1^2 - \sum_{n_1}\frac{||x_{n_1} - \mu_1||^2}{2\sigma_1^2} + n_2\log w_2 - \frac{n_2}{2}\log 2\pi \sigma_2^2 - \sum_{n_2}\frac{||x_{n_2} - \mu_2||^2}{2\sigma_2^2}
\end{multline}
 
The parameters $w_1, w_2, \mu_1$, $\mu_2$, $\sigma_1$ and $\sigma_2$ are unknown. We use the maximum likelihood estimates as a plug-in for each:
\begin{align*}
    \hat{w}_1 &= n_1 / N \\
    \hat{\mu}_1 &= \frac{1}{n_1}\sum_{n < s} x_{n} \\
    \hat{\sigma_1} &= \frac{1}{n_1}\sum_{n < s}||x_{n} - \hat{\mu}_1||^2,
\end{align*}
with the parameters for $j=2$ defined equivalently. 
% sample means to estimate the means i.e $\hat{\mu_1} = \frac{1}{n_1}\sum_{n_1}x_{n_1}$ and $\hat{\mu_2} = \frac{1}{n_2}\sum_{n_2}x_{n_2}$.We use the sample variances to estimate the variances. i.e $\hat{\sigma_1} = \frac{1}{n_1}\sum_{n_1}||x_{n_1} - \hat{\mu_1}||^2$ and $\hat{\sigma_2} = \frac{1}{n_2}\sum_{n_2}||x_{n_2} - \hat{\mu_2}||^2$.  
Thus the right hand side of Equation \eqref{eq2} can be expressed as

\begin{align}\label{eq3}
 n_1\log w_1 - \frac{n_1}{2}\log 2\pi \hat{\sigma}_1^2 - \frac{n_1}{2}  + n_2\log w_2 - \frac{n_2}{2}\log 2\pi \hat{\sigma}_2^2 -\frac{n_2}{2}
\end{align}

But in the BIC formula, we need to compute the negative log likelihood.
\begin{align}\label{eq3}
    -\log \prod_n P(x_n;\hat{\mu}, \hat{\sigma}^2, \hat{w}) =  -n_1\log \hat{w}_1 + \frac{n_1}{2}\log 2\pi \hat{\sigma}_1^2 + \frac{n_1}{2}  - n_2\log \hat{w}_2 + \frac{n_2}{2}\log 2\pi \hat{\sigma}_2^2 + \frac{n_2}{2}.
\end{align}

\section{Supplemental Figures}

\begin{figure}%[H]
    \centering %\offinterlineskip
        \includegraphics[width=\textwidth]{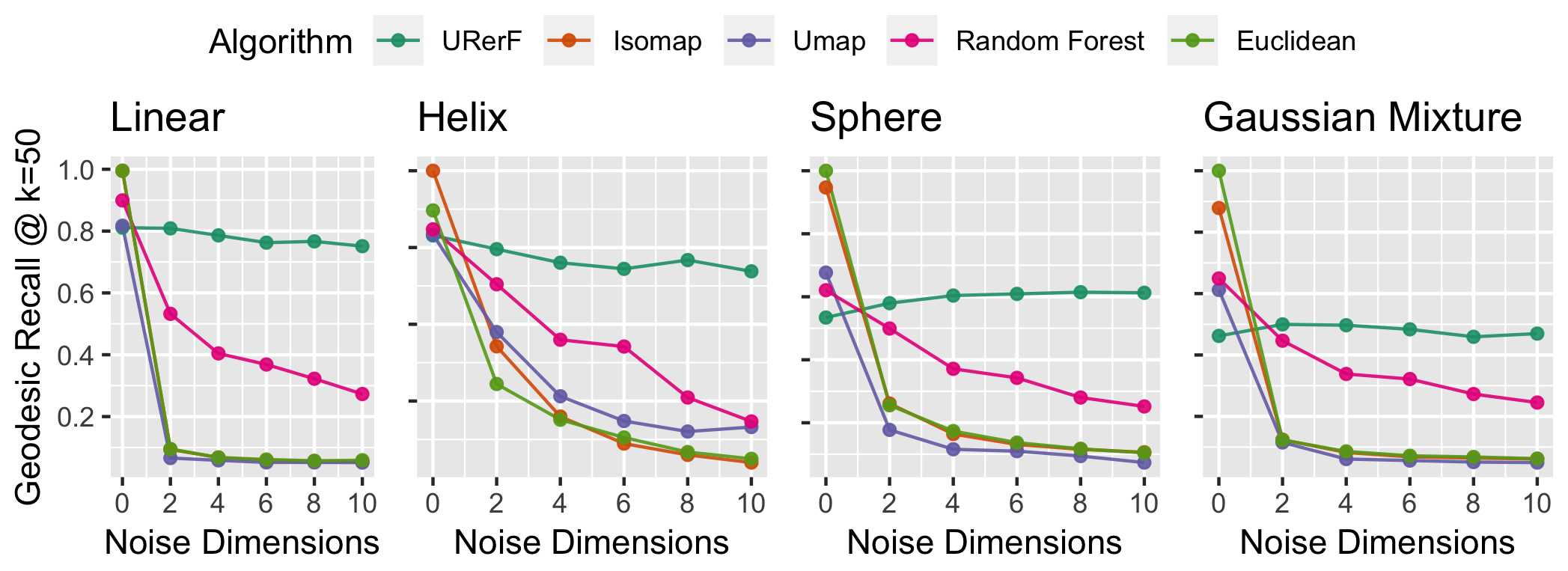} 
        \caption{Geodesic precision at k=50 with varying noise dimension  $d'$ from 2 to 10, using $N = 1000$ samples, and  $p = 3$  number of non-noisy dimensions.  URerF is robust to adding noise dimensions while the other algorithms deteriorate to chance in performance.
        }
        \label{fig2} 
        % TODO remove "noise dimensions" everywhere except the far left panels, and replace with "# of noise dimensions". make panels all equal width (they are not currently.
\end{figure}

\end{document}